# Detecting Qualia
## in Natural and Artificial Agents

**Roman V. Yampolskiy**
Computer Engineering and Computer Science
Speed School of Engineering
University of Louisville
roman.yampolskiy@louisville.edu

*"The greatest obstacle to discovery is not ignorance—it is the illusion of knowledge."*
Daniel J. Boorstin

*"Consciousness is the one thing in this universe that cannot be an illusion."*
Sam Harris

**Abstract**
The Hard Problem of consciousness has been dismissed as an illusion. By showing that computers are capable of experiencing, we show that they are at least rudimentarily conscious with potential to eventually reach superconsciousness. The main contribution of the paper is a test for confirming certain subjective experiences in a tested agent. We follow with analysis of benefits and problems with conscious machines and implications of such capability on future of computing, machine rights and artificial intelligence safety.

## 1. Introduction to the Problem of Consciousness

One of the deepest and most interesting questions ever considered is the nature of consciousness. An explanation for what consciousness is, how it is produced, how to measure it or at least detect it [1] would help us to understand who we are, how we perceive the universe and other beings in it, and maybe even comprehend the meaning of life. As we embark on the quest to create intelligent machines, the importance of understanding consciousness takes on the additional fundamental role and engineering thoroughness. As presence of consciousness is taken to be the primary reason for granting many rights and ethical consideration [2], its full understanding will drastically change how we treat our mind children and perhaps how they treat us.

Initially the question of consciousness was broad and ill-defined encompassing problems related to intelligence, information processing, free will, self-awareness, essence of life and many others. With better understanding of brain architecture and progress in artificial intelligence and cognitive science many easy sub-problems of consciousness have been successfully addressed [3] and multiple neural correlates of consciousness identified [4]. However, some fundamental questions remain as poignant as ever: What is it like to be bat? [5], What is it like to be a brain simulation? [10], etc. In other words, what is it like to be a particular type of an agent [6-9]? What it feels like to be one? Why do we feel something at all? Why red doesn't sound like a bell [11]? What red looks like [12]? What is it like to see with your tongue [13]? In other words, we are talking about experiencing what it is like to be in a particular state. Block [14] calls it *Phenomenal* or P-consciousness to distinguish it from *Access* or A-consciousness. David Chalmers managed to distill away non-essential components of consciousness and suggested that explaining qualia (what it feels like to experience something) and why we feel in the first place as opposed to being philosophical zombies [15] is the Hard Problem of consciousness [3]:

"The really hard problem of consciousness is the problem of *experience*. When we think and perceive, there is a whir of information processing, but there is also a subjective aspect. As Nagel (1974) has put it, there is *something it is like* to be a conscious organism. This subjective aspect is experience. When we see, for example, we *experience* visual sensations: the felt quality of redness, the experience of dark and light, the quality of depth in a visual field. Other experiences go along with perception in different modalities: the sound of a clarinet, the smell of mothballs. Then there are bodily sensations from pains to orgasms; mental images that are conjured up internally; the felt quality of emotion; and the experience of a stream of conscious thought. What unites all of these states is that there is something it is like to be in them. All of them are states of experience." [3]. " … [A]n organism is conscious if there is something it is like to be that organism, and a mental state is conscious if there is something it is like to be in that state. Sometimes terms such as "phenomenal consciousness" and "qualia" are also used here, but I find it more natural to speak of "conscious experience" or simply "experience."" [3]

Daniel Dennet [16] and others [17] have argued that in fact there is no Hard Problem and that what we perceive as consciousness is just an illusion like many others, an explanation explored by scholars of illusionism [18-21]. Over the years a significant amount of evidence has been collected all affirming that much of what we experience is not real [22], including visual [23-25], auditory [26], tactile [27], gustational [28], olfactory [29], culture specific [30] and many other types of illusions [31]. An illusion is a discrepancy between agent's awareness and some stimulus [32]. Illusions can be defined as stimuli which produce a surprising percept in the experiencing agent [33] or as a difference between perception and reality [34]. As we make our case mostly by relying on Visual Illusions in this paper, we include the following definition from García-Garibay et al.: "Visual illusions are sensory percepts that can't be explained completely from the observed image but that arise from the internal workings of the visual system." [35].

Overall, examples of illusions may include: impossible objects [36], blind spot [37], paradoxes (Zeno's [38], mathematical/logical illusions [39]), quantum illusions [40], mirages [41], art [42, 43], Rorschach tests [44], acquired taste [45], reading jumbled letters [46], forced perspective [47], gestaltism [48], priming [49], stereograms [50], delusion boxes [51], temporal illusions [52], constellations [53], illusion within an illusion [54], world [55], Déjà Vu [56], reversing goggles [57], rainbows [58], virtual worlds [59], and wireheading [60]. It seems that illusions are not exceptions, they are the norm in our world, an idea which was rediscovered through the ages [61-63].

Moreover, if we take a broader definition and include experiences of different states of consciousness, we can add: dreams (including lucid dreams [64] and nightmares [65]), hallucinations [66], delusions [67], drug induced states [68], phantom pains [69], religious experiences [70], self [71] (homunculus [72]), cognitive biases [73], mental disorders, invisible disabilities and perception variations (Dissociative identity disorder [74], Schizophrenia [75, 76], Synesthesia [77], Simultanagnosia [78], Autism [79], Ideasthesia [80], Asperger's [81], Apophenia [82], Aphantasia [83], Prosopagnosia [84] – all could be reclassified as issues with "correctly" experiencing illusions), Pareidolia [85], ironic processes [86], emotions (love, hate) [87], feelings (hunger, pain, pleasure) [88], body transfer [89], out of body experiences [90], sensory substitution [91], novel senses [92], and many others.

Differences between what is traditionally considered to be an illusion and what we included can be explained by how frequently we experience them. For example, the sky looks different depending on the time of day, amount of Sun or the angle you are experiencing it from, but we don't consider it to be an illusion because we experience it so frequently. Essentially, everything can be considered to be an illusion, the difference is that some stimuli are very common while others are completely novel to us, like a piece of great art, see for example [42]. This makes us think that if we experience something many times it is real, but if we see something for the first time it must be an illusion.

At the extreme, we can treat every experience is an illusion in which some state of atomic particles in the universe is perceived as either a blue sky, or a beautiful poem or a hot plate or a conscious agent. This realization is particularly obvious in the case of digital computers, which are machines capable of extrapolating all the world's objects from strings of binary digits. Isn't experiencing a face in a bunch of zeroes and ones a great illusion, in particular while another machine experiences a melody on the same set of inputs ([93], p44.)?

Likewise, neurodiverse individuals may experience the world in very different ways, just consider color blindness [94] as example of same inputs being experienced differently by diverse types of human agents. In fact, we suggest that most mental disorders can be better understood as problems with certain aspects of generating, sustaining or analyzing illusions [75]. Similarly, with animals, studies show that many are capable of experiencing same illusions as people [95-98], while also experiencing our world in a very different way [99]. Historically, we have been greatly underestimating consciousness of animals [100], and it is likely that now we are doing it to intelligent machines.

What it feels like to be a particular type of agent in a given situation depends on the hardware/software/state of the agent and stimulation being provided by the environment. As the qualia represent the bedrock of consciousness, we can formally define a conscious agent as one capable of experiencing at least some broadly defined illusions. To more formally illustrate this we can represent the agent and its inputs as two shares employed in visual cryptography [101], depending on the composition of the agent the input may end up producing a diametrically opposite experience [102, 103]. Consequently, consciousness is an ability to experience, and we can state two ways in which illusions, and consciousness may interact to produce a conscious agent:

- An agent is real and is experiencing an illusion. This explains qualia and the agent itself is real.
- An agent is real and is having an illusion in which some other agent experiences an illusion. Self-identifying with such an agent creates self-consciousness. A sequence of such episodes corresponds to a stream of consciousness and the illusionary agent itself is not real. You are an illusion experiencing an illusion.

## 2. Test for Detecting Qualia

Illusions provide a tool [104, 105], which makes it possible to sneak a peek into the mind of another agent and determine that an agent has in fact experienced an illusion. The approach is similar to non-interactive CAPTCHAs, in which some information is encoded in a CAPTCHA challenge [106-110] and it is only by solving the CAPTCHA correctly that the agent is able to obtain information necessary to act intelligently in the world, without having to explicitly self-report its internal state [111-114]. With illusions, it is possible to set up a test in which it is only by experiencing an illusion that the agent is able to enter into a certain internal state, which we can say it experiences. It is not enough to know that something is an illusion. For example, with a classical face/vase illusion [115] an agent who was previously not exposed to that challenge, could be asked to report what two interpretations for the image it sees and if the answer matches that of a human experiencing that illusion the agent must also be experiencing the illusion, but perhaps in a different way.

Our proposal represents a variant of a Turing Test [116, 117] but with emphasis not on behavior or knowledge but on experiences, feelings and internal states. In related research, Schweizer [118] has proposed a Total Turing Test for Qualia (Q3T), which is a variant of Turing Test for a robot with sensors and questions concentrated on experiences such as: how do you find that wine? Schneider and Turner have proposed a behavior based AI consciousness test, which looks at whether the synthetic mind has an experience-based understanding of the way it feels to be conscious as demonstrated by an agent "talking" about consciousness related concepts such as afterlife or soul [119].

What we describe is an empirical test for presence of some subjective experiences. The test is probabilistic but successive different variants of the test can be used to obtain any desired level of confidence. If a collaborating agent fails a particular instance of the test it doesn't mean that the agent doesn't have qualia, but passing an instance of the test should increase our belief that the agent has experiences in proportion to the chance of guessing correct answer for that particular variant of the test. As qualia are agent type (hardware) specific (human, specie, machine, etc.) it would be easiest for us to design a human-compatible qualia test, but in principle, it is possible to test for any type of qualia, even the ones that humans don't experience themselves. Obviously, having some qualia doesn't mean ability to experience them all. While what we propose is a binary detector test for some qualia, it is possible to design specific variants for extracting particular properties of qualia experience such as color, depth, size, etc. The easiest way to demonstrate construction of our test is by converting famous visual illusions into instances of our test questions as seen in Figure 1. Essentially we present our subject with an illusion and ask it a multiple choice question about the illusionary experience, such as: how many black dots do you see? How many curved lines are in the image? Which of the following affects do you observe? It is important to only test subjects with tests they have not experienced before and information about which is not readily available. Ideally a new test question should be prepared every time to prevent the subject from cheating. A variant of the test may ask open ended questions such as: please describe what you see. In that case, a description could be compared to that produced by a conscious agent, but this is less formal and opens the door for subjective interpretation of submitted responses. Ideally, we want to be able to automatically design novel illusions with complex information encoded in them as experiences.

| **Horizontal lines are:** | **Orange circles are:** | **Horizontal stripe is:** |
|---|---|---|
| 1) Not in the image<br>2) Crooked<br>3) Straight<br>4) Red | 1) Left one is bigger<br>2) Right one is bigger<br>3) They are the same size<br>4) Not in the image | 1) Solid<br>2) Spectrum of gray<br>3) Not in the image<br>4) Crooked |
| 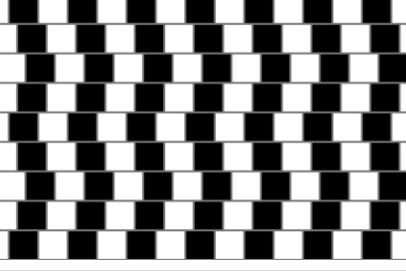 | 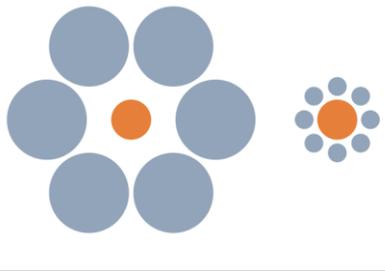 | 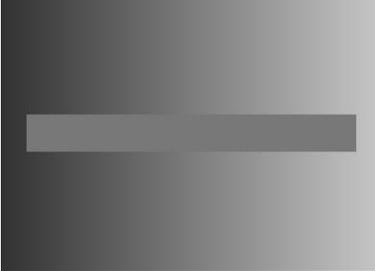 |
| By Fibonacci - Own work, CC BY-SA 3.0,<br>https://commons.wikimedia.org/w/index.php?curid=1788689 | Public Domain,<br>https://commons.wikimedia.org/w/index.php?curid=828098 | By Dodek - Own work, CC BY-SA 3.0,<br>https://commons.wikimedia.org/w/index.php?curid=1529278 |

Figure 1. *Visual Illusions presented as tests*.

We anticipate a number of possible objections to the validity of our test and its underlying theory:

- **Qualia experienced by the test subject may not be the same as experienced by the test designer.**
  Yes, we are not claiming that they are identical experiences; we are simply showing that an agent had some subjective experiences, which was previously not possible. If sufficiently different, such alternative experiences would not result in passing of the test.
- **The system may simply have knowledge of the human mental model and predict what a human would experience on similar stimulus.**
  If a system has an internal human (or some other) model which it simulates on presented stimuli and that generates experiences, it is the same as the whole system having experiences.
- **Agent may correctly guess answers to the test or lie about what it experiences.**
  Yes, for a particular test question, but the test can be given as many times as necessary to establish statistical significance.
- **The theory makes no predictions.**
  We predict that computers built to emulate the human brain will experience progressively more illusions without being explicitly programmed to do so, in particular the ones typically experienced by people.
- Turing addressed a number of relevant objections in his seminar paper on computing machinery [120].

## 3. Computers can Experience Illusions and so are Conscious

Majority of scholars studying illusionism are philosophers, but a lot of relevant work comes from psychology [121], cognitive science [122] and more recently computer science, artificial intelligence, machine learning and more particularly Artificial Neural Network research. It is this interdisciplinary nature of consciousness research which we think is most likely to produce successful and testable theories, such as the theory presented in this paper, to solve the Hard problem.

In the previous section, we have established that consciousness is fundamentally based on an ability to experience, for example illusions. Recent work with artificially intelligent systems suggests that computers also experience illusions and in a similar way to people, providing support for the Principle of Organizational Invariance [3] aka substrate independence [55]. For example, Zeman et al. [123, 124] and Garcia-Garibay et al. [35] report on a neural networks capable of experiencing Müller-Lyer illusion and multiple researchers [125-128] have performed experiments in which computer models were used to study visual illusions, including teaching computers to experience geometric illusions [125, 129, 130], brightness illusions [131, 132] and color constancy illusions [133]. In related research, Nguyen et al. found that NN perceive certain random noise images as meaningful with very high confidence [134]. Those NN were not explicitly designed to perceive illusions but they do so as a byproduct of the computations they perform. The field of Adversarial Neural Networks is largely about designing illusions for such intelligent systems [135, 136] with obvious parallels to inputs known to fool human agents intentionally [137] or unintentionally [138]. Early work on artificial Neural Networks, likewise provides evidence for experiences similar to near death hallucinations [139, 140] (based on so-called "virtual inputs" or "canonical hallucination" or "neural forgery" [141]), dreaming [142, 143], and impact from brain damage [144, 145].

Zeman [34] reviews history of research on perception of illusions by computer models and summarizes the state-of-the-art in such research: "Historically, artificial models existed that did not contain multiple layers but were still able to demonstrate illusory bias. These models were able to produce output similar to human behaviour when presented with illusory figures, either by emulating the filtering operations of cells [127, 146] or by analysing statistics in the environment [126, 147-149]. However, these models were deterministic, non-hierarchical systems that did not involve any feature learning. It was not until Brown and Friston (2012) [150] that hierarchical systems were first considered as candidates for modelling illusions, even though the authors omitted important details of the model's architecture, such as the number of layers they recruited. … So to summarise, illusions can manifest in artificial systems that are both hierarchical and capable of learning. Whether these networks rely on exposure to the same images that we see during training, or on filtering mechanisms that are based on similar neural operations, they produce a consistent and repeatable illusory bias. In terms of Marr's (1982) [151] levels of description …, it appears that illusions can manifest at the hardware level [148, 149] and at the algorithmic/representational level [123, 127, 146]."

"By dissociating our sensory percepts from the physical characteristics of a stimulus, visual illusions provide neuroscientists with a unique opportunity to study the neuronal mechanisms underlying … sensory experiences" [35]. Not surprisingly artificial neural networks just like their natural counterparts are subject to similar analysis. From this, we have to conclude that even today's simple AIs, as they experience specific types of illusions, are rudimentary conscious.

General intelligence is what humans have and we are capable of perceiving many different types of complex illusions. As AIs become more adept at experiencing complex and perhaps multisensory illusions they will eventually reach and then surpass our capability in this domain producing multiple parallel streams of superconsciousness [152], even if their architecture or sensors are not inspired by the human brain. Such superintelligent and superconscious systems could justifiably see us as barely intelligent and weakly conscious, and could probably control amount of consciousness they had, within some range. Google deep dream art [153] gives us some idea on what it's like to be a modern deep neural network and can be experienced in immersive 3D via the Hallucination Machine [154]. Olah et al. provide a detailed neuron/layer visual analysis of what is being perceived by an artificial neural network [155].

### 3.1 Qualia Computing

If we can consistently induce qualia in computational agents, it should be possible to use such phenomena to perform computation. If we can encode information in illusions, certain agents can experience them or their combinations to perform computation, including artificially intelligent agents capable of controlling their illusions. Illusions are particularly great to represent superpositions of states (similar to quantum computing), which collapse once a particular view of the illusion is chosen by the experiencing agent [156]. You can only experience one interpretation of an illusion at a time, just like in Quantum physics you can only know location or speed of a particle at the same time - well known conjugate pairs [157]. Famous examples of logical paradoxes can be seen as useful[1] for super-compressed data storage [158, 159] and hyper-computation [160]. Qualia may also be useful in explaining decisions produced by deep NN, with the last layer efficiently representing qualia-like states derived from low-level stimuli by lower level neurons. Finally, qualia based visualization and graphics are a very interesting area of investigation, with the human model giving us an example of visual thinking and lucid dreaming.

### 4. Purpose of Consciousness

While many scientific theories, such as biocentrism [161] or some interpretations of quantum physics [162, 163], see consciousness as a focal element of their models, the purpose of being able to experience remains elusive. In fact, even measurement or detection of consciousness remains an open research area [1]. In this section, we review and elaborate on some explanations for what consciousness does. Many explanations have been suggested, including but certainly not limited to [20]: error monitoring [164], an inner eye [165], saving us from danger [166], later error detection [167], pramodular response [168] and to seem mysterious [169].

We can start by considering the evolutionary origins of qualia from the very first, probably accidental, state of matter, that experienced something, all the way to general illusion experiences of modern humans. The argument is that consciousness evolved because accurately representing reality is less important than agents' fitness for survival and agents who saw the world of illusions had higher fitness, as they ignored irrelevant and complicated minutia of the world [170]. It seems that processing real world is computationally expensive and simplifying illusions allow improvements in efficiency of decision-making leading to higher survival rates. For example, we can treat feelings as heuristic shortcuts to calculating precise utility. Additionally, as we argue in this paper, experiencing something allows one to obtain knowledge about that experience, which is not available to someone not experiencing the same qualia. Therefore, a conscious agent would be able to perform in ways a philosophical zombie would not be able to act, which is particularly important in the world full of illusions such as ours.

---

[1] Kolmogorov complexity is also not computable, but very useful.

Next, we can look at the value of consciousness in knowledge acquisition and learning. A major obstacle to the successful development of AI systems, has been what is called the Symbol Grounding problem [171]. Trying to explain to a computer one symbol in terms of others does not lead to understanding. For example saying that "mother" is a female parent is no different than saying that $x = 7y$, and $y = 18k$ and so on. This is similar to a person looking up an unfamiliar word in a foreign language dictionary and essentially ending up with circular definitions of unfamiliar terms. We think, that qualia are used (at least in humans) to break out of this vicious cycle and to permit definitions of words/symbols in terms of qualia. In "*How Helen Keller used syntactic semantics to escape from a Chinese Room*", Rappaport [172] gives a great example of a human attempting to solve the grounding problem and argues that syntactic semantics are sufficient to resolve it. We argue that it is experiencing the feeling of running water on her hands was what permitted Hellen Keller to map sign language sign for water to the relevant qualia and to begin to understand.

Similarly, we see much of the language acquisition process as mapping of novel qualia to words. By extension, this mapping permits us to explain understanding and limits to transfer of tacit knowledge. Illusion disambiguation can play a part in what gives us an illusion of free will and the stream of consciousness may be nothing more than sequential illusion processing. Finally, it would not be surprising if some implicit real-world inputs produced experience of qualia behind some observed precognition results [173]. In the future, we suspect a major application of consciousness will be in the field of Qualia Computing as described in the so-named section of this paper.

**4.1 Qualia Engineering**
While a grand purpose of life remains elusive and is unlikely to be discovered, it is easy to see that many people attempt to live their lives in a way, which allows them to maximally explore and experience novel stimuli: foods, smells, etc. Experiencing new qualia by transferring our consciousness between different substrates, what Loosemore refers to as Qualia Surfing [174], may represent the next level in novelty seeking. As our understanding and ability to detect and elicit particular qualia in specific agents improves, qualia engineering will become an important component of the entertainment industry. Research in other fields such as: intellectology [175], (and in particular artimetrics [176, 177], and designometry [178]), consciousness [179] and artificial intelligence [180] will also be impacted.

People designing optical illusions, movie directors and book authors are some of the people in the business of making us experience, but they do so as an art form. Qualia engineers and qualia designers will attempt to formally and scientifically answer such questions as: How to detect and measure qualia? What is the simplest possible qualia? How to build complex qualia from simple ones? What makes some qualia more pleasant? Can minds be constructed with maximally pleasing qualia in a systematic and automated way [175]? Can this lead to abolition of suffering [181]? Do limits exist to complexity of qualia, or can the whole universe be treated as single input? Can we create new feelings and emotions? How would integration of novel sensors expand our qualia repertoire? What qualia are available to other agents but not to humans? Can qualia be "translated" to other mediums? What types of verifiers and observers experience particular types of qualia? How to generate novel qualia in an algorithmic/systematic way? Is it ethical to create unpleasant qualia? Can agents learn to swap qualia between different stimuli (pleasure for pain)? How to optimally represent, store and communicate qualia, including across different substrates [55]? How to design an agent, which experiences particular qualia on the given input? How much influence does an agent have over its own illusions? How much plasticity does the human brain have for switching stimuli streams and learning to experience data from new sensors? How similar are qualia among similarly designed but not identical agents? What, if any, is the connection between meditation and qualia? Can computers mediate? How do random inputs such as café chatter [182] stimulate production of novel qualia? How can qualia be classified into different types, for example feelings? Which computations produce particular qualia?

## 5. Consciousness and Artificial Intelligence

Traditionally, AI researchers ignored consciousness as non-scientific and concentrated on making their machines capable and beneficial. One famous exception is Hofstadter who observed and analyzed deep connections between illusions and artificial intelligence [183]. If an option to make conscious machines presents itself to AI researchers, it would raise a number of important questions, which should be addressed early on. It seems that making machines conscious may make them more relatable and human like and so produce better consumer products, domestic and sex robots and more genuine conversation partners. Of course, a system simply simulating such behaviors without actually experiencing anything could be just as good. If we define physical pain as an unpleasant sensory illusion and emotional pain as an illusion of an unpleasant feeling, pain and pleasure become accessible controls to the experimenter. Ability to provide reward and punishment for software agents capable of experiencing pleasure and pain may assist in the training of such agents [184].

Potential impact from making AI conscious includes change in the status of AI from mere useful software to a sentient agent with corresponding rights and ethical treatment standards. This is likely to lead to civil rights for AI and disenfranchisement of human voters [185, 186]. In general, ethics of designing sentient beings are not well established and it is cruel to create sentient agents for certain uses, such as menial jobs, servitude or designed obsolescence. It is an experiment which would be unlikely to be approved by any research ethics board [187]. Such agents may be subject to abuse as they would be capable of experiencing pain and torture, potentially increasing the overall amount of suffering in the universe [188]. If in the process of modeling or simulating conscious beings, experiment negatively affects modeled entities this can be seen as mind crime [189].

With regards to AI safety [190-195], since it would be possible for agents to experience pain and pleasure it will open a number of new pathways for dangerous behavior. Consciousness may make AIs more volatile or unpredictable impacting overall safety and stability of such systems [119]. Possibility of ransomware with conscious artificial hostages comes to mind as well as blackmail and threats against AI system. Better understanding of consciousness by AI itself may also allow superintelligent machines to create new types of attacks on people. Certain illusions can be seen as an equivalent of adversarial inputs for human agents, see Figure 2. Subliminal stimuli [196] which confuse people are well known and some stimuli are even capable of inducing harmful internal states such as epileptic seizures [197, 198] or incapacitation [199]. With latest research showing, that even a single pixel modification is sufficient to fool neural networks [200], the full scope of the attack surface against human agents remains an unknown unknown.

Manual attempts to attack a human cognitive model are well known [201-203]. Future research combining evolutionary algorithms or adversarial neural networks with direct feedback from detailed scans of human brains is likely to produce some novel examples of adversarial human inputs, leading to new types of informational hazards [204]. Taken to the extreme, whole adversarial worlds may be created to confuse us [55]. Nature provides many examples of adversarial inputs in plants and animals, known as mimicry [205]. Human adversarial inputs designed by superintelligent machines would represent a new type of AI risk, which has not been previously analyzed and with no natural or synthetic safety mechanisms available to defend us against such an attack.

One very dangerous outcome from integration of consciousness into AI is a possibility that a superintelligent system will become a negative utilitarian and an anti-natalist [188] and in an attempt to rid the world of suffering will not only kill all life forms, but will also destroy all AIs and will finally self-destruct as it is itself conscious and so subject to the same analysis and conclusions. This would result in a universe free of suffering but also free of any consciousness. Consequently, it is important to establish guidelines and review boards [206] for any research which is geared at producing conscious agents [207]. AI itself should be designed to be corrigible [208] and to report any emergent un-programmed capabilities, such as qualia, to the designers.

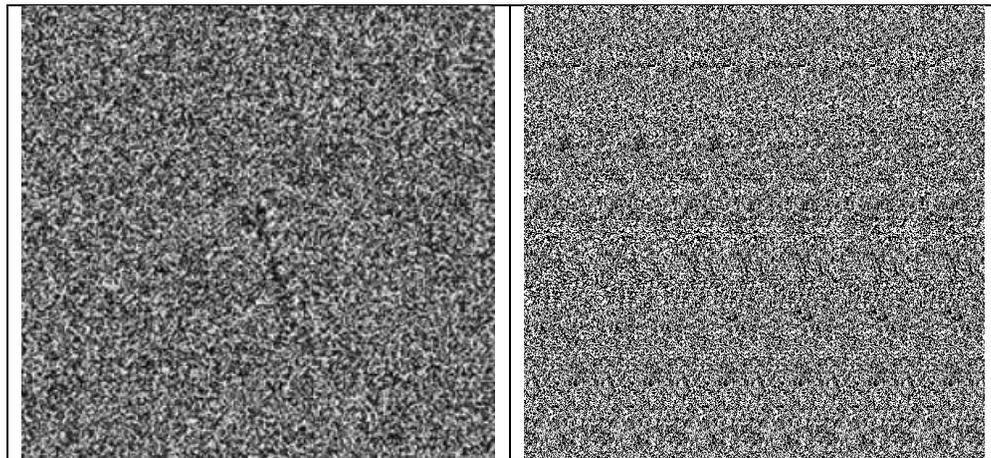

Figure 2: *Left - Cheetah in the noise is seen by some Deep Neural Networks (based on [134]); Right - Spaceship in the stereogram is seen by some people.*

## 6. Conclusions and Conjectures

In this paper, we described a reductionist theory for appearance of qualia in agents based on a fully materialistic explanation for subjective states of mind, an attempt at a solution to the Hard Problem of consciousness. We defined a test for detecting experiences and showed how computers can be made conscious in terms of having qualia. Finally, we looked at implications of being able to detect and generate qualia in artificial intelligence. Should our test indicate presence of complex qualia in software or animals certain protections and rights would be appropriate to grant to such agents. Experimental results, we surveyed in this paper, have been predicted by others as evidence of consciousness in machines, for example Dehaene et al. state: "We contend that a machine endowed with [global information availability and self-monitoring] … may even experience the same perceptual illusions as humans" [209].

Subjective experiences called qualia are a side effect of computing, unintentionally produced while information is being processed, similar to generation of heat [210], noise [211], or electromagnetic radiation [212] and is just as unintentional. Others have expressed similar intuitions: "The cognitive algorithms we use are the way the world feels." ([213] p. 889.) or "consciousness is the way information feels when being processed." [214] or "empirical evidence is compatible with the possibility that consciousness arises from nothing more than specific computations" [209]. Qualia arise as a result of processing of stimuli caused by agglomeration of properties, unique peculiarities [215] and errors in agent's architecture, software, memories, learned algorithms, sensors, inputs, environment and other factors comprising extended cognition [216] of an agent [217]. In fact, Zeman [34] points out the difficulty of telling if a given system experiences an error or an illusion. If every computation produces side effect of qualia, computational functionalism [218] trivially reduces to panpsychism [219].

As qualia are fully dependent on a makeup of a particular agent it is not surprising that they capture what it is like to be that agent. Agents, which share certain similarities in their makeup (like most people), may share certain subsets of qualia, but different agents will experience different qualia on the same inputs. An illusion is a discrepancy between agent's awareness and some stimulus [32]. In contrast, consciousness is an ability to experience a sustained self-referential multimodal illusion based on an ability to perceive qualia. Every experience is an illusion, what we call optical illusions are meta illusions, there are also meta-meta-illusions and self-referential illusions. It is an illusion of "I" or self which produces self-awareness, with "I" as an implied agent experiencing all the illusions, an illusion of an illusion navigator.

It is interesting to view the process of learning in the context of this paper, with illusions as a primary pattern of interest for all agents. We can say that babies and other untrained neural networks are learning to experience illusions, particularly in the context of their trainers' culture/common sense [30]. Consequently, a successful agent will learn to map certain inputs to certain illusions while sharing that mapping with other similarly constructed observers. We can say that the common space of illusions/culture as seen by such agents becomes their "real world" or meme [220] sphere. Some supporting evidence for this conclusion comes from observing that amount of sleep in children is proportionate to the average amount of learning they perform for that age group. Younger babies need the most sleep, perhaps because they can learn quicker by practicing to experience in the safe world of dreams (a type of illusion) a skill they then transfer to the real world. Failure to learn to perceive illusions and experience qualia may result in a number of mental disorders.

There seems to be a fundamental connection between intelligence, consciousness and liveliness beyond the fact that all three are notoriously difficult to define. We believe that ability to experience is directly proportionate to one's intelligence and that such intelligent and conscious agents are necessarily alive to the same degree. As all three come in degrees, it is likely that they have gradually evolved together. Modern narrow AIs are very low in general intelligence and so are also very low in their ability to experience or their perceived liveness. Higher primates have significant (but not complete) general intelligence and so can experience complex stimuli and are very much alive. Future machines will be superintelligent, superconscious and by extension alive!

Fundamental "particles" from which our personal world is constructed are illusions, which we experience and in the process create the universe, as we know it. Experiencing a pattern which is not really there (let's call such an illusory element "illusination"), like appearing white spaces in an illusion [221], is just like experiencing self-awareness; where is it stored? Since each conscious agent perceives a unique personal universe, their agglomeration gives rise to the multiverse. We may be living in a simulation, but from our point of view we are not living in a virtual reality [222], we are living in an illusion of reality, and maybe we can learn to decide which reality to create. The "Reality" provides us with an infinite set of inputs from which every conceivable universe can be experienced and in that sense, every universe exists. We can conclude that the universe is in the mind of the agent experiencing it - the ultimate qualia, even if we are just brains in a vat, to us an experience is worth a 1000 pictures. It is not a delusion that we are just experiencers of illusions. Brain is an illusion experiencing machine not a pattern recognition machine. As we age, our wetware changes and so we become different agents and experience different illusion, our identity changes but in a continuous manner. To paraphrase Descartes: I experience, therefore I am conscious!

## Acknowledgements

The author is grateful to Elon Musk and the Future of Life Institute and to Jaan Tallinn and Effective Altruism Ventures for partially funding his work on AI Safety. The author is thankful to Yana Feygin for proofreading a draft of this paper and to Ian Goodfellow for helpful recommendations of relevant literature.